\documentclass[10pt,twocolumn,letterpaper]{article}

\usepackage{cvpr}              
\definecolor{cvprblue}{rgb}{0.21,0.49,0.74}
\usepackage[pagebackref,breaklinks,colorlinks,allcolors=cvprblue]{hyperref}
\usepackage{amsmath}
\usepackage{amssymb}
\usepackage{multirow}
\usepackage{booktabs,subcaption}
\usepackage{tcolorbox}
\usepackage{pifont}
\newcommand{\cmark}{\ding{51}}%

\def\paperID{} 
\def\confName{CVPR}
\def\confYear{2026}

\title{RI-Mamba: Rotation-Invariant Mamba for Robust Text-to-Shape Retrieval}

\author{
    Khanh Nguyen, Dasith de Silva Edirimuni, Ghulam Mubashar Hassan \& Ajmal Mian \\
    The University of Western Australia \\
    \texttt{duykhanh.nguyen@research.uwa.edu.au} \\ 
    \texttt{\{dasith.desilva,ghulam.hassan,ajmal.mian\}@uwa.edu.au} \\
}

\begin{document}
\maketitle
\begin{abstract}
3D assets have rapidly expanded in quantity and diversity 
due to the growing popularity of virtual reality and gaming. As a result, text-to-shape retrieval has become essential in facilitating intuitive search within large repositories. However, existing methods require canonical poses and support few object categories, limiting their real-world applicability where objects can belong to diverse classes and appear in random orientations. To address this challenge, we propose RI-Mamba, the first rotation-invariant state-space model for point clouds. RI-Mamba defines global and local reference frames to disentangle pose from geometry and uses Hilbert sorting to construct token sequences with meaningful geometric structure while maintaining rotation invariance. 
We further introduce a novel strategy to compute orientational embeddings and reintegrate them via feature-wise linear modulation, effectively recovering spatial context and enhancing model expressiveness. Our strategy is inherently compatible with state-space models and operates in linear time.
To scale up retrieval, we adopt cross-modal contrastive learning with automated triplet generation, allowing training on diverse datasets without manual annotation. Extensive experiments demonstrate RI-Mamba's superior representational capacity and robustness, achieving state-of-the-art performance on the OmniObject3D benchmark across more than 200 object categories under arbitrary orientations. Our code will be made available at \url{https://github.com/ndkhanh360/RI-Mamba.git}.
\end{abstract}    
\section{Introduction}
The increasing availability of 3D modeling technologies and advancements in computing hardware have led to a rapid expansion of 3D assets. This growth is further driven by the rising demand in virtual/augmented reality and gaming. As large-scale 3D repositories continue to expand, efficient and intuitive retrieval methods become essential for practical applications, such as selecting a 3D furniture model for interior design or retrieving an object to put into a virtual scene. Natural language offers a user-friendly interface for such tasks, motivating the development of text-to-shape retrieval methods \cite{y2seq2seq,tricolo,sca3d}.

\begin{figure}[t]
    \centering
    \includegraphics[width=0.9\linewidth]{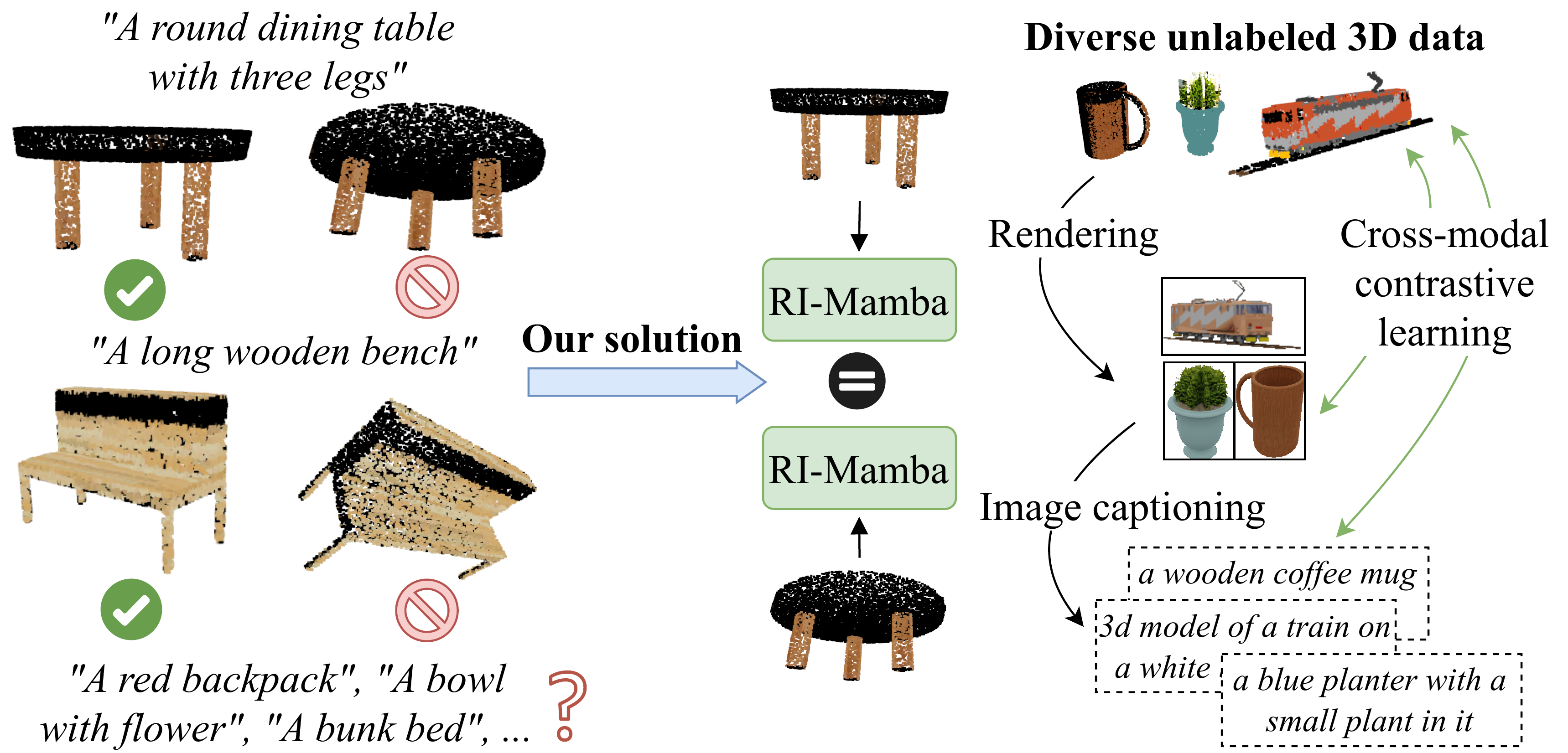}
    \caption{Existing text-to-3D retrieval methods require human annotations and pre-aligned shapes, limiting them to a narrow set of object classes and restricting them to canonical poses. 
    Our proposed RI-Mamba is designed to be rotation invariant and is trained via cross-modal contrastive learning on diverse 3D assets without manual annotations to enable retrieval across a wide range of object categories under arbitrary orientations.}
    \label{fig:motivation}
\end{figure}

However, existing works face two key limitations that hinder their practical use: \textit{i)} they are restricted to a small set of predefined object categories, 
and \textit{ii)} they assume all 3D objects are in their canonical poses. These constraints arise because current approaches rely heavily on human descriptions for supervision, and state-of-the-art (SOTA) models require fine-grained part segmentation labels to achieve accurate text-shape alignment. Such annotations are only available in small-scale curated datasets with objects aligned to predefined poses, making it difficult to scale to more varied objects with arbitrary orientations often found in online 3D shape repositories or large-scale datasets \cite{sketchfab,objaverse_xl,objaverse}. Consequently, existing methods struggle to retrieve randomly rotated objects or generalize to novel categories, as illustrated in Fig.~\ref{fig:motivation}.

To overcome the first limitation, we adopt a simple yet effective cross-modal contrastive learning framework to align 3D representations with image and text embeddings. This approach has demonstrated impressive performance in zero-shot 3D recognition \cite{ulip,openshape,tamm,duomamba}, and exhibits strong potential for retrieval tasks. However, its effectiveness for fine-grained text-to-3D retrieval remains underexplored. Tricolo \cite{tricolo} simplifies complex retrieval pipelines \cite{parts2words,y2seq2seq} with contrastive learning and achieves promising results, but still relies on human annotations and is limited to only 13 ShapeNet categories \cite{shapenet}. In contrast, we leverage automated data generation based on the rendering process, text prompts, and image captioning to eliminate the need for manual labeling. This enables us to significantly scale up training and allow text-to-shape retrieval across more than 200 object categories. 

To address the pose variation problem, we introduce RI-Mamba, a novel rotation-invariant (RI) architecture for point clouds. To our knowledge, it is the first Mamba-based model that achieves invariance under arbitrary SO(3) rotations. Although several RI networks have been proposed for point cloud recognition~\cite{riconv++,pose_aware_conv,locotrans,ritransformer}, they are designed and trained specifically for single-modal tasks. These architectures often impose strict constraints~\cite{se3_cnn,spherical_fractal_cnn,tfn}, exhibit limited expressiveness~\cite{clusternet,riframework,riconv++}, or incur high computational cost~\cite{ritransformer}. 
In particular, the SOTA RI-Transformer~\cite{ritransformer} achieves the best performance but suffers from quadratic complexity due to its attention mechanism, limiting its scalability to cross-modal training. In comparison, RI-Mamba leverages linear-time state-space models~\cite{mamba} to ensure computational efficiency while maintaining high performance. 

However, achieving rotation invariance with state-space models remains an open problem, requiring careful design to ensure that both patch embeddings and their positions within the sequence are rotation-invariant. RI-Mamba addresses this challenge through a principled and effective design. 
For robust embeddings, it establishes a local reference frame (LRF) for each patch and aligns the patch into its LRF, effectively factoring out pose information from the geometric content. To define the token order, our RI serialization module introduces a global reference frame (GRF) and applies a Hilbert space-filling curve to sort patch centers within the GRF, ensuring a consistent order under rotation. 
While this approach guarantees rotation invariance, aligning patches to LRFs inherently discards their pose information, a critical factor that often limits the expressiveness of RI models. 
To overcome this problem, we propose an orientational embedding that encodes each patch's orientation relative to the entire object. Unlike prior quadratic \textit{pairwise} methods that require relative orientations between all patch pairs \cite{ritransformer}, our \textit{patchwise} approach achieves linear-time complexity, which is a requirement for state-space models. 
The pose information is reintegrated via feature-wise linear modulation, allowing the model to adapt point features to spatial context and thereby enhancing its representational capacity.

To summarize, our main contributions are:
\begin{enumerate}
    \item We introduce RI-Mamba, the first rotation-invariant architecture based on state-space models, achieving competitive performance while reducing computational cost compared to the state-of-the-art RI-Transformer.

    \item We design patchwise orientational embeddings to be compatible with Mamba.
    Unlike prior pairwise orientations, our embeddings are computed in linear time and integrate seamlessly with state-space models.
    
    \item We introduce text-to-shape training without manual annotation by adopting automated data generation and cross-modal contrastive learning. This enables scaled-up training and retrieval on 200+ object categories.
    
    \item We establish the first benchmark for text-to-shape retrieval on diverse objects with arbitrary poses, enabling realistic and comprehensive assessment. 
\end{enumerate}

\section{Related Works}
\begin{figure*}[t]
    \centering
    \includegraphics[width=0.98\linewidth]{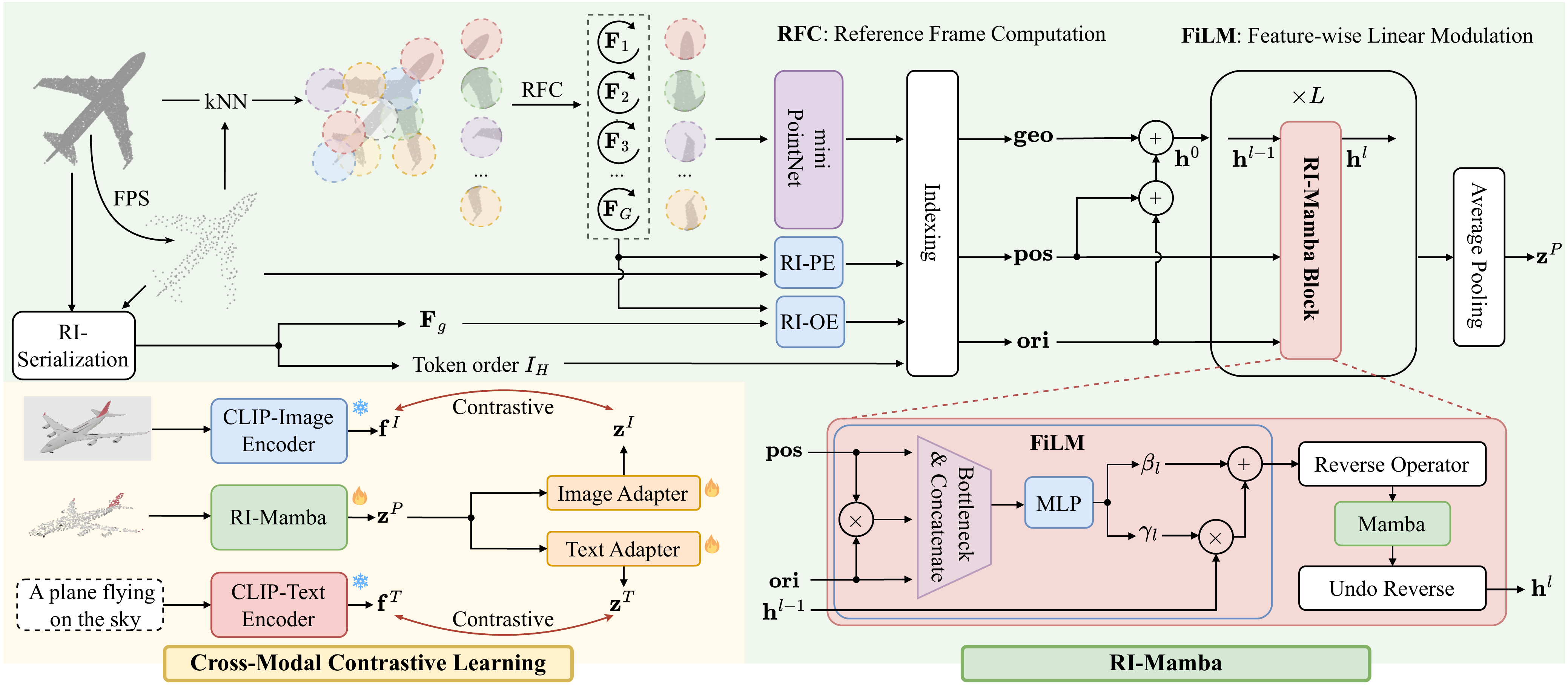}
    \caption{Overview of our method. Given a point cloud, we form local patches using FPS and kNN. RI-Serialization and RFC align patches and define a RI token order. RI geometric, positional, and orientational embeddings are extracted and fed into RI-Mamba blocks, which model long-range relationships via FiLM, reverse operators, and Mamba modules \textbf{(green)}. The final 3D feature is aligned with CLIP's image and text embeddings via cross-modal contrastive learning \textbf{(yellow)}.}
    \label{fig:overview}
\vspace{-3mm}    
\end{figure*}
\noindent \textbf{Zero-Shot 3D Recognition.}
ULIP~\cite{ulip} is 
a pioneering work to unify text, image, and 3D shape representations via cross-modal contrastive learning. It leverages rendered images of 3D meshes and caption templates to align a 3D encoder with CLIP’s text and image embeddings~\cite{clip}, enabling zero-shot 3D recognition. Subsequent works extend this by scaling up the training data with automated triplet generation~\cite{openshape}, increasing the capacity of the 3D encoder~\cite{uni3d}, and reducing the domain gap between rendered and real-world images~\cite{tamm}. Most recently, DuoMamba \cite{duomamba} was proposed to address the quadratic complexity of transformers using state-space models, along with occlusion-aware pretraining to improve real-world recognition.
Although these methods demonstrate strong zero-shot performance, their aligned embedding spaces have not been quantitatively evaluated for fine-grained text-to-3D retrieval. Moreover, none of them addresses the challenge of unconstrained object poses commonly found in large-scale 3D repositories. 

\vspace{1mm}
\noindent \textbf{Text-to-Shape Retrieval.}
\citeauthor{text2shape} \cite{text2shape} pioneered the combination of metric learning and learning-by-association to align 3D shape and natural language embeddings extracted by 3D-CNN and GRU encoders. They also introduced the first text-to-3D retrieval benchmark named Text2Shape, which includes only table and chair objects from ShapeNet~\cite{shapenet}. Y$^2$Seq2Seq~\cite{y2seq2seq} addresses the low-resolution issue of 3D-CNN with a view-based approach that jointly reconstructs and predicts multi-view images and captions. Parts2Words \cite{parts2words} incorporates part segmentation annotations and leverages optimal transport to match object parts with words, improving 3D-text similarity computation and retrieval accuracy. COM3D~\cite{com3d} further exploits cross-view correspondences and applies SRT~\cite{srt} to enrich 3D features. SCA3D~\cite{sca3d} utilizes part-level information and a large vision-language model~\cite{llava} to generate detailed descriptions for object parts, which are then assembled to augment training data. However, the focus on limited object categories (only chairs and tables) and reliance on costly part-level annotations restricts the scalability of these methods to real-world diverse data.
Tricolo~\cite{tricolo} explores cross-modal contrastive learning as a way to eliminate the need for part segmentation labels, achieving competitive results and enabling retrieval over 13 object categories. Nevertheless, it still relies on human-annotated captions, which limits further scalability. Moreover, these methods all prioritize the training procedure for retrieval without focusing on designing a model that can handle arbitrary object poses commonly encountered in real-world retrieval conditions. 

\noindent \textbf{Rotation-Invariant Models for Point Clouds.}
Several RI methods are built on spherical CNNs~\cite{se3_cnn, spherical_fractal_cnn}, which achieve invariance through discretization and spherical projection. Although effective, these operations can result in the loss of fine-grained geometric details and limit flexibility for diverse 3D tasks. Tensor Field Networks~\cite{tfn} offer a family of SO(3)-equivariant models based on group theory but impose strict architectural constraints and suffer from high computational overhead. Other approaches~\cite{clusternet, riframework, riconv++} design handcrafted rotation-invariant features, such as distances and angles, to replace Cartesian coordinates as input. However, such methods typically discard orientation information, reducing their expressiveness and degrading their performance. Pose-aware convolution \cite{pose_aware_conv} was proposed to mitigate this by assigning adaptive weights based on the relative poses between points. However, this method requires costly pairwise computations. Similarly, LocoTrans~\cite{locotrans} attempts to preserve pose information via a relative pose recovery module, but this necessitates an additional equivariant branch, doubling the overall computation. More recently, RI-Transformer~\cite{ritransformer} leverages geometry and orientation disentanglement to ensure rotation robustness. While it achieves state-of-the-art performance, its attention-based architecture incurs a high computational cost. 

In this paper, we propose the first Mamba-based RI architecture to efficiently address the challenge of arbitrary poses in real-world retrieval. We further adopt automated data generation and cross-modal contrastive learning to eliminate the need for manual annotation and scale up text-to-3D retrieval.
\section{Proposed Method}
\textbf{Overview.} 
Given an input point cloud with $N$ points, denoted as $\mathbf{P}_0 \in \mathbb{R}^{N \times 3}$ (optionally with color information $\mathbf{C}_0 \in \mathbb{R}^{N \times 3}$), we first apply {Farthest Point Sampling} (FPS) to select $G$ representative center points $\mathbf{P} \in \mathbb{R}^{G \times 3}$. Around the $i$-th center, we gather $k$ neighboring points using k-Nearest Neighbors (kNN), forming a local patch with relative coordinates with respect to the center $\mathbf{p}_i \in \mathbb{R}^{k \times 3}$.
To disentangle geometry from pose and ensure local rotation invariance, we compute a Local Reference Frame (LRF) $\mathbf{F}_i$ for each patch and align the local points within its LRF. We then apply our Rotation-Invariant Serialization (RI-Serialization) to convert the unordered set of patches into a geometrically consistent sequence that is robust to random global rotations.
Each patch is encoded using a mini-PointNet~\cite{pointnet} to extract a geometric feature vector $\mathbf{geo}_i$ with $C$ channels. We further compute positional and orientational embeddings ($\mathbf{pos}_i, \mathbf{ori}_i$) to help recover information lost during previous projections. These embeddings are then fed into a series of $L$ {RI-Mamba blocks} to model long-range dependencies and high-level geometry.
Finally, we perform average pooling to get a single point cloud feature $\mathbf{z}^P \in \mathbb{R}^C$ and align it with CLIP's image and text embeddings through cross-modal contrastive learning. Fig.~\ref{fig:overview} illustrates our pipeline. 

\subsection{Reference Frame Computation (RFC)}
\label{subsec:rfc}
Given any set of points $\mathbf{X} \in \mathbb{R}^{n \times 3}$, we interpret it as a rotated version of an underlying canonical point set $\hat{\mathbf{X}}$ defined in a reference frame $\mathbf{F} \in \mathbb{R}^{3 \times 3}$. The canonical point set $\hat{\mathbf{X}}$ captures the intrinsic geometry, while the rotation encapsulated in $\mathbf{F}$ encodes the orientation of $\mathbf{X}$. To decouple geometry from pose and ensure robustness to arbitrary rotations, we compute the reference frame $\mathbf{F}$ and project the original points into this frame as:
\begin{align}
    \label{eq:rot_norm}
    \mathbf{F} = \mathrm{RFC}(\mathbf{X}), \quad \hat{\mathbf{X}} = \mathbf{X} \mathbf{F}^\top.
\end{align}

Following prior works~\cite{local_keypoint,ritransformer}, we estimate $\mathbf{F}$ using Principal Component Analysis (PCA). While PCA provides orthogonal axes that reflect major directions of variance in the point set, it is inherently ambiguous in the sign of each axis; i.e., both directions along an axis are equally valid. This ambiguity can cause inconsistencies during sequential processing.
To address this, we perform reference frame disambiguation by projecting the point set onto each axis and selecting the direction along which the majority of points lie as the positive direction. This simple yet effective strategy ensures consistent and deterministic results.

We apply this procedure to each local patch $\mathbf{p}_i$ to obtain its LRF $\mathbf{F}_i$, and align the patch as $\hat{\mathbf{p}}_i = \mathbf{p}_i \mathbf{F}_i^\top$, yielding a normalized representation that captures local geometry independent of rotation.

\subsection{Rotation-Invariant Point Serialization}
\label{subsec:serialization}
Our model is built on Mamba~\cite{mamba}, a unidirectional state-space model where each token's update depends on its position in the input sequence. Consequently, it is critical to serialize the unordered set of point patches into a 1D sequence that preserves meaningful geometric structure to enable effective information propagation across tokens~\cite{duomamba}. Moreover, this serialization must be robust to the arbitrary pose of the input object to maintain the model's rotation invariance.

To this end, we first compute a global reference frame (GRF) that captures the orientation of the entire object, following the procedure described in Sec.~\ref{subsec:rfc}. We then project the patch centers onto the GRF to obtain their canonical coordinates, which are invariant to rotations. To impose a spatially meaningful order on the patches, we sort the projected centers using a Hilbert space-filling curve~\cite{hilbert_curve}, which preserves spatial locality by mapping nearby 3D points to adjacent positions in the 1D sequence.  Specifically: 
\begin{align}
    \label{eq:ri_serialization}
    \mathbf{F}_g = \mathrm{RFC}(\mathbf{P}_0), \quad 
    \hat{\mathbf{P}} = \mathbf{P} \mathbf{F}_g^\top, \quad 
    I_H = \mathrm{Hilbert}(\hat{\mathbf{P}}).
\end{align}

To verify the rotation-invariant property of this serialization, consider applying a rotation $\mathbf{R}$ to the input point cloud: $\mathbf{P}_0^r = \mathbf{P}_0 \mathbf{R}$, which results in rotated patch centers $\mathbf{P}^r = \mathbf{P} \mathbf{R}$ and a rotated GRF $\mathbf{F}_g^r = \mathbf{F}_g \mathbf{R}$. Substituting into Eq.~\ref{eq:ri_serialization}, we obtain:
\begin{align}
    \label{eq:ri_proof}
    \hat{\mathbf{P}}^r = \mathbf{P}^r (\mathbf{F}_g^r)^\top 
    = \mathbf{P} \mathbf{R} (\mathbf{F}_g \mathbf{R})^\top 
    = \mathbf{P} \mathbf{F}_g^\top = \hat{\mathbf{P}},
\end{align}
which shows that the projected centers in the global reference frame remain unchanged under any rotation $\mathbf{R}$. As a result, the Hilbert index $I_H$ is also preserved. This guarantees that our serialization is invariant to the orientation of the input, enabling consistent and robust sequence modeling across arbitrary poses.

\subsection{Rotation-Invariant Patch Embeddings}
\label{subsec:ri_embeddings}
Following previous Transformer and Mamba-based point cloud models \cite{pointmamba,pointbert,duomamba}, we represent each point patch with a token embedding that captures local geometric structure and a positional embedding that encodes its spatial location within the point cloud. In our RI-Mamba, the geometric embedding for the $i$-th patch is extracted from its aligned coordinates using a mini-PointNet:
\begin{align}
    \label{eq:pointnet}
    \mathbf{geo}_i = \mathrm{PointNet}(\hat{\mathbf{p}}_i) = \mathrm{PointNet}(\mathbf{p}_i \mathbf{F}_i^\top).
\end{align}

Similarly, we compute the positional embedding by projecting the patch center $\mathbf{P}_i$ into its LRF and passing it through an MLP:
\begin{align}
    \label{eq:pos_emb}
    \mathbf{pos}_i = \mathrm{MLP}(\hat{\mathbf{P}}_i) = \mathrm{MLP}(\mathbf{P}_i \mathbf{F}_i^\top).
\end{align}
When a random rotation is applied to the input point cloud, both $\mathbf{p}_i$ and $\mathbf{P}_i$ rotate accordingly, but their projections into the LRF ($\hat{\mathbf{p}}_i$ and $\hat{\mathbf{P}}_i$) remain unchanged (see Eq.~\ref{eq:ri_proof}). As a result, the inputs to both $\mathrm{PointNet}$ and $\mathrm{MLP}$ in Eq.~\ref{eq:pointnet} and Eq.~\ref{eq:pos_emb} are preserved, ensuring that the geometric and positional embeddings are invariant to global rotation.

\paragraph{Linear-time Orientational Embedding.}
While the above process guarantees rotation robustness, it inherently discards the orientation information of each patch due to the canonical projections. Previous studies~\cite{ritransformer,pose_aware_conv,locotrans} have shown that such loss of pose information can reduce model expressiveness and harm the performance.
To address this, we introduce an orientational embedding that recovers patch-level pose information. A naive solution is to compute an embedding from each LRF using an MLP. However, since LRFs themselves rotate with the input, such embeddings would not be rotation invariant. Another approach, proposed by RI-Transformer~\cite{ritransformer}, models pairwise relative orientations between patches to preserve invariance, but this incurs quadratic computation and is incompatible with the unidirectional processing of state-space models.

Instead, we propose a more efficient strategy that models the relative orientation of each patch with respect to the global pose of the object as follows:
\begin{align}
    \label{eq:ori_emb}
    \mathbf{ori}_i = \mathrm{MLP}(\mathbf{F}_i^\top \mathbf{F}_g).
\end{align}
Since both $\mathbf{F}_i$ and $\mathbf{F}_g$ are affected in the same way by an arbitrary global rotation, their relative pose $\mathbf{F}_i^\top \mathbf{F}_g$ remains unchanged, ensuring that $\mathbf{ori}_i$ is rotation-invariant. 
By computing the pose of each patch relative to the GRF, we obtain an independent orientational embedding for each patch, making them compatible with state-space models. This also eliminates explicit pairwise comparisons with quadratic complexity as in RI-Transformer, and allows the model to capture inter-patch orientation through sequential processing implicitly, resulting in lower computational cost. 

Finally, all patch embeddings ($\mathbf{geo}_i$, $\mathbf{pos}_i$, and $\mathbf{ori}_i$) are sorted using the rotation-invariant order $I_H$ from Sec.~\ref{subsec:serialization}, forming a sequence that is passed to a stack of RI-Mamba blocks for higher-level feature extraction.

\subsection{RI-Mamba Blocks}
An overview of the design is illustrated in Fig.~\ref{fig:overview}. The $l$-th RI-Mamba block receives the hidden states $\mathbf{h}_{l-1} \in \mathbb{R}^C$ (i.e. the transformed point tokens from the previous layer) along with the positional and orientational embeddings. It processes these inputs through a Feature-wise Linear Modulation (FiLM) module, a Reverse Operator, and a Mamba module to propagate information between the patches and extract long-range geometric relationships. 

\paragraph{Feature-wise Linear Modulation (FiLM).}  
To recover information potentially lost during the previous patch tokenization, we incorporate positional and orientational context into the hidden states $\mathbf{h}_l$ at each layer $l$ using FiLM mechanism~\cite{film}.
This module learns channel-wise scale and shift parameters that modulate the hidden states, allowing dynamic adjustment of point features based on their spatial context. In particular, we use a bottleneck layer to reduce the dimension of $\mathbf{pos}$, $\mathbf{ori}$, and their element-wise product $\mathbf{pos} \odot \mathbf{ori}$, then concatenate them and use an MLP to learn the scale and bias $\gamma_l, \beta_l \in \mathbb{R}^C$ to modify the hidden states as $\mathbf{h}'_l = \gamma_l \cdot \mathbf{h}_{l-1} + \beta_l$.

\paragraph{Reverse Operator for Bidirectional Scanning.}  
Our model leverages Mamba modules~\cite{mamba}  for linear-time modeling of inter-patch relationships and high-level geometry. However,  Mamba is inherently causal and only processes the input sequence in a single direction. While this is suitable for sequential data like language or audio, it is suboptimal for point clouds where aggregating neighbors' information from all directions is essential~\cite{duomamba}.

Therefore, we employ a reverse operator to enable bidirectional scanning of the input sequence. In particular, we alternate the processing direction across layers: odd-numbered layers scan the sequence from left to right, while even-numbered layers scan from right to left. This simple yet effective mechanism facilitates richer contextual flows, improving performance without extra computation.

\subsection{Cross-Modal Contrastive Learning}
\label{subsec:contrastive_learning}
To align 3D representations with the rich image and text features produced by CLIP~\cite{clip}, we train our RI-Mamba model following the cross-modal contrastive learning framework from TAMM~\cite{tamm}. In particular, after averaging the output of the final RI-Mamba block to obtain a global point cloud embedding $\mathbf{z}^P$, we decouple this feature into visual attributes $\mathbf{z}^I$ and semantic latents $\mathbf{z}^T$ via two adapters. They are then aligned with CLIP's image $\mathbf{f}^I$ and text embeddings $\mathbf{f}^T$, which are extracted from the corresponding image and text in the training triplets to minimize:
\begin{equation}
\label{eq:totoal_loss}
    \mathbb{L} = \mathcal{L}^{P \leftrightarrow I} + \mathcal{L}^{P \leftrightarrow T},
\end{equation}
where $\mathcal{L}^{P \leftrightarrow M}$ denotes the InfoNCE loss~\cite{infonce} between the two modalities $P$ (point cloud) and $M \in \{I, T\}$ (image or text), defined as:
\begin{equation}
\scriptsize 
\label{eq:info_nce}
\resizebox{0.48\textwidth}{!}{$
 - \frac{1}{2B} \sum_{i=1}^{B} \left( \log\frac{\exp\left(\mathbf{z}_i^{M} \cdot \mathbf{f}_i^{M} / \tau \right)}{\sum_{j=1}^{B} \exp\left(\mathbf{z}_i^{M} \cdot \mathbf{f}_j^{M} / \tau \right)} + \log\frac{\exp\left(\mathbf{f}_i^{M} \cdot \mathbf{z}_i^{M} / \tau \right)}{\sum_{j=1}^{B} \exp\left(\mathbf{f}_i^{M} \cdot \mathbf{z}_j^{M} / \tau \right)} \right),
 $}
\end{equation}
with $B$ is the batch size and $\tau$ is a temperature hyperparameter that controls the sharpness of the distribution.
\section{Experiments}
\label{sec:experiment}
\label{subsec:rimamba}
\paragraph{Experimental Setup.} 
We conduct two primary text-to-shape retrieval experiments under \textit{supervised} and \textit{zero-shot} settings, considering both canonical poses and random rotations (referred to as SO(3)). 

In line with prior works~\cite{ritransformer,locotrans}, the SO(3) evaluation results are reported from a single run, using identical uniformly sampled rotations for all models to ensure fair and consistent comparison.

In supervised retrieval, we evaluate our method against previous approaches designed to retrieve table and chair objects from Text2Shape dataset~\cite{text2shape}, which contains 11,498 and 1,434 shapes for training and testing, respectively. 
This experiment explores whether simple cross-modal contrastive learning can compete with previous \textit{training procedures} that rely on complex losses and part-level supervision. 

For the zero-shot setting, we compare the performance of RI-Mamba with other \textit{architectures} when trained using the same cross-modal contrastive learning procedure introduced in TAMM~\cite{tamm} (see Sec.~\ref{subsec:contrastive_learning}). Our comparison includes non-RI models originally used in zero-shot 3D recognition (PointBERT~\cite{pointbert} and DuoMamba~\cite{duomamba}) as well as RI architectures (LocoTrans~\cite{locotrans} and RI-Transformer~\cite{ritransformer}). 
We evaluate text-to-shape retrieval on the OmniObject3D test set~\cite{omniobject3d}, which comprises 1,251 shapes across 214 categories, each paired with a detailed text description from a human expert.
This experiment establishes the first benchmark that resembles real-world retrieval scenarios and provides a quantitative assessment for existing zero-shot models.
For comprehensive evaluation, we further report results on zero-shot classification and 3D-to-3D retrieval. Complete experimental and implementation details are provided in the supplementary.

\paragraph{Pretraining Data for Zero-Shot Experiments.} 
We pretrain 3D encoders using point cloud-image-text triplets provided by OpenShape~\cite{openshape}. These triplets are automatically generated via rendering and image captioning from 3D meshes sourced from ShapeNet~\cite{shapenet}, ABO~\cite{abo}, 3D-FUTURE~\cite{3dfuture}, and Objaverse-LVIS~\cite{objaverse}. To evaluate the effect of dataset scale, we compare performance when pretraining on ShapeNet alone ($\sim$52K objects) versus on the ensemble of all four datasets ($\sim$123K objects).

\subsection{Supervised Text-to-Shape Retrieval}
\label{sec:supervised_text2pcd}
We first compare our method with existing text-to-shape retrieval approaches, including Y$^2$Seq2Seq~\cite{y2seq2seq}, COM3D~\cite{com3d}, TriCoLo~\cite{tricolo}, Parts2Words~\cite{parts2words}, and SCA3D~\cite{sca3d}. Evaluations are conducted on the Text2Shape~\cite{text2shape} dataset, which contains 3D models of tables and chairs. Given a natural language query, the goal is to retrieve the correct 3D shape from a database. We use Recall Rate at $k$ (RR@$k$) and Normalized Discounted Cumulative Gain (NDCG) as evaluation metrics. RR@k measures the frequency with which the ground-truth shape appears among the top-$k$ retrieved candidates, while NDCG evaluates the quality of the ranking relative to an ideal ordering. All models are trained on the Text2Shape training split, and results on the test set are reported in Tab.~\ref{tab:text2shape}.
\begin{table}[t]
\huge    
\centering
    \resizebox{0.48\textwidth}{!}{%
    \begin{tabular}{lcccccc}
    \toprule
    \multirow{2}{*}{Method} & \multicolumn{3}{c}{Text2Shape} & \multicolumn{3}{c}{Text2Shape-SO(3)} \\ \cline{2-7} 
                   & \rule{0pt}{2.2ex}RR@1  & RR@5  & NDCG@5 & RR@1  & RR@5  & NDCG@5 \\ 
    \midrule
    \midrule

    Y$^2$Seq2Seq & 2.93 & 9.23 & 6.05 & -  & -  & -   \\
    COM3D & 5.64 & 18.50 & 12.09 & -  & -  & -   \\  
    TriCoLo & 12.22 & 32.23 & 22.46 & -  & -  & -   \\ 
    \midrule
    COM3D$^\dagger$ & 13.12 & 33.48 & 23.89 & -  & -  & -   \\ 
    Parts2Words$^\dagger$ & 12.72 & 32.98 & 23.13 & 1.68 & 5.46  & 3.57    \\ 
SCA3D$^\dagger$ & \underline{13.74} & \textbf{35.11} & \textbf{24.58} & \underline{2.24} & \underline{6.59} & \underline{4.46}   \\   
    \midrule
    RI-Mamba (ours) & \textbf{13.87} & \underline{34.57} & \underline{24.55} & \textbf{13.20} & \textbf{33.62}	& \textbf{23.65}   \\
    \bottomrule
    \multicolumn{7}{l}{($^\dagger$ denotes results obtained using part segmentation labels for training)}
    \end{tabular}
    }
\caption{Supervised text-to-3D results on Text2Shape.}
\label{tab:text2shape}
\end{table}

Our method outperforms all approaches that do not rely on part-level annotations and is competitive with SOTA methods that leverage extra part labels for more accurate text-shape matching. Under the rotation setting, RI-Mamba significantly surpasses all other methods. This is because prior works primarily focus on the training procedure and overlook the robustness of their models, resulting in substantial performance drops when database objects are randomly rotated.
In contrast, our carefully designed RI-Mamba achieves both high expressiveness and impressive robustness, as demonstrated by its strong performance across all evaluation settings and metrics.

\begin{figure}[t]
    \centering
    \includegraphics[width=\linewidth]{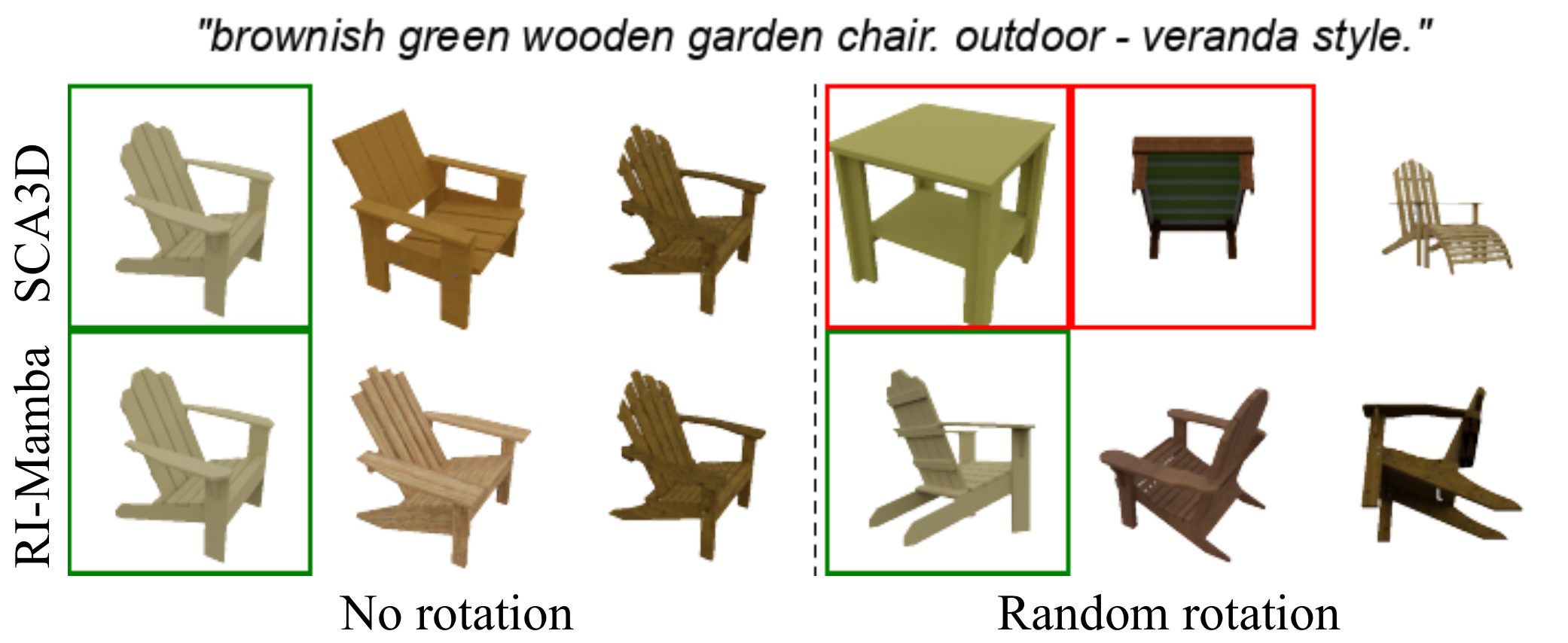}
    \caption{Retrieval results on Text2Shape dataset. Under random rotation, RI-Mamba maintains robust performance while SCA3D fails to retrieve the correct chair (green box) and includes irrelevant tables (red boxes) in top 3 candidates.}
    \label{fig:text2pcd_retrieval}
\end{figure}

\paragraph{Visualization.} We compare the retrieval results of SCA3D and RI-Mamba in Fig.~\ref{fig:text2pcd_retrieval}. When 3D shapes are in canonical poses, both methods correctly retrieve the target object. However, under arbitrary orientations, SCA3D fails to retrieve the correct chair described in the query and incorrectly includes tables in the top 3 candidates. In contrast, RI-Mamba consistently retrieves only relevant chairs that closely match the textual description. This demonstrates the challenges of real-world retrieval and highlights the strong practical applicability of our proposed model. More visualizations are in the supplementary.

\subsection{Zero-Shot Text-to-Shape Retrieval}
\label{subsec:zeroshot_text2shape}
To simulate a real-world deployment scenario, where a trained model is expected to perform text-to-shape retrieval on an unseen dataset with diverse object categories, we conduct zero-shot text-to-3D retrieval on OmniObject3D, which covers 214 object classes. The 3D encoders are pretrained on ShapeNet or Ensemble data and evaluated on the OmniObject3D test set without any fine-tuning.

\begin{table}[t]
\Huge 
\centering
    \resizebox{0.48\textwidth}{!}{%
    \begin{tabular}{lccccccc}
    \toprule
    \multirow{2}{*}{Data} & \multirow{2}{*}{Method} & \multicolumn{3}{c}{Omni3D} & \multicolumn{3}{c}{Omni3D-SO(3)} \\ \cline{3-8} 
             &                & \rule{0pt}{2.2ex}RR@1  & RR@5  & NDCG@5 & RR@1  & RR@5  & NDCG@5 \\ 
    \midrule
    \midrule
    \multirow{5}{*}{\rotatebox{90}{ShapeNet}} & PointBERT-TAMM & 7.11  & 19.66 & 13.61  & 2.96  & 8.07  & 5.53   \\
             & DuoMamba      &   \textbf{8.79}    &   \underline{22.86}    &    \underline{16.16}    & 3.28       & 11.19      & 7.25       \\
             \cline{2-8}\rule{0pt}{2.2ex}
             & LocoTrans      &   2.96    &  12.79     &    7.82    &   2.80    &   11.67    &    7.27    \\
             & RITransformer  & 3.92  & 12.47 & 8.23   & \underline{3.76}  & \underline{12.39} & \underline{8.12}   \\
             & RI-Mamba (ours)       & \underline{8.39}  & \textbf{24.70}  & \textbf{16.80}   & \textbf{9.27 } & \textbf{23.74} & \textbf{16.70}   \\ 
    \midrule
    \multirow{5}{*}{\rotatebox{90}{Ensemble}} & PointBERT-TAMM & 15.67 & 41.57 & 28.84  & 8.93  & 23.18 & 16.18  \\
             & DuoMamba      &   \underline{15.82}    &  \underline{41.89}     &   \underline{29.28}     &   7.83    &  25.98     &    17.09    \\    
             \cline{2-8}\rule{0pt}{2.2ex}
             & LocoTrans      &  7.75   &  28.70     &  18.46      &   8.39    &   28.14    &   18.40     \\
             & RITransformer  & 9.67  & 31.73 & 20.89  & \underline{10.39} & \underline{31.65} & \underline{21.24}  \\
             & RI-Mamba (ours)       & \textbf{19.02} & \textbf{49.08} & \textbf{34.39}  & \textbf{19.34} & \textbf{47.16} & \textbf{33.76} \\ 
    \bottomrule
    \end{tabular}
    }
\caption{Zero-shot text-to-3D results on OmniObject3D.}
\vspace{-3mm}
\label{tab:zeroshot_text2shape}
\end{table}

Tab.~\ref{tab:zeroshot_text2shape} shows that non-RI models experience significant performance drops under random rotations. In contrast, RI methods maintain more stable performance under rotation but often suffer from limited expressiveness, obtaining lower accuracy than non-RI approaches on aligned point clouds. Notably, our RI-Mamba achieves the best performance across nearly all settings, demonstrating impressive representational power and robustness.
An important observation is that all methods benefit substantially from scaling up the training data.
This highlights the effectiveness of cross-modal contrastive learning and automated triplet generation in leveraging large-scale 3D data, offering a scalable solution to improve real-world applicability.

\subsection{Zero-Shot 3D-to-3D Retrieval}
We further evaluate our model’s generalization on zero-shot 3D-to-3D retrieval, where both the query and database objects are point clouds from an unseen dataset. Given a query, the model must retrieve all database objects from the same category. We randomly split the OmniObject3D dataset into database (80\%) and query (20\%) sets, with 4,571 and 1,251 shapes, respectively. Following standard practice in single-modal retrieval, we adopt mean Average Precision (mAP) as the evaluation metric. To assess robustness to rotation, we consider four settings: both query and database in canonical pose (I/I), only the query rotated (I/SO(3)), only the database rotated (SO(3)/I), and both rotated (SO(3)/SO(3)).

\begin{table}[h]
\centering
    \resizebox{0.44\textwidth}{!}{%
    \begin{tabular}{lccccc}
    \toprule
    Data & Method & \multicolumn{1}{c}{I/I} & \multicolumn{1}{c}{I/SO(3)} & \multicolumn{1}{c}{SO(3)/I} & \multicolumn{1}{c}{SO(3)/SO(3)} \\ 
    \midrule
    \midrule
    \multirow{5}{*}{\rotatebox{90}{ShapeNet}} & PointBERT-TAMM & \textbf{43.12} & 12.77 & 12.88 & 15.42 \\
             & DuoMamba      &   \underline{41.16}    & 12.18      & 12.38       & 15.69       \\ 
             \cline{2-6}\rule{0pt}{2.2ex}
             & LocoTrans      &   33.39 & 33.43 & 33.26 & 33.15       \\
             & RITransformer  & 36.39 & \underline{36.19} & \underline{36.47} & \underline{36.26} \\
             & RI-Mamba  (ours)      & {39.51} & \textbf{39.33} & \textbf{39.60}  & \textbf{39.40}  \\
    \midrule
    \multirow{5}{*}{\rotatebox{90}{Ensemble}}  & PointBERT-TAMM & \textbf{49.84} & 21.91 & 21.74 & 24.42 \\
             & DuoMamba      &   46.81 & 19.93 & 21.02 & 25.30 \\
             \cline{2-6}\rule{0pt}{2.2ex}
             & LocoTrans      &   39.83 & 39.54 & 39.72 & 39.59       \\
             & RITransformer  & 43.57 & \underline{43.63} & \underline{43.75} & \underline{43.78} \\
             & RI-Mamba (ours)      & \underline{47.58} & \textbf{47.56} & \textbf{47.48} & \textbf{47.50} \\ 
             \bottomrule
    \end{tabular}
    }
\caption{Zero-shot 3D-to-3D mAP on OmniObject3D.}
\label{tab:pcd2pcd}
\end{table}

As shown in Tab.~\ref{tab:pcd2pcd}, standard models such as PointBERT and DuoMamba perform well in the I/I setting due to their strong representational capacity. However, their performance degrades significantly under three random rotation settings. RI methods maintain stable performance across all rotation conditions but generally underperform in the I/I scenario. 
Our proposed RI-Mamba achieves the highest mAP across all rotation settings while remaining competitive in the canonical pose condition. This demonstrates its strong balance between representational power and rotation robustness, making it well-suited for real-world 3D retrieval applications where fixed object orientation is not guaranteed.

\subsection{Zero-Shot 3D Classification}
We further assess model performance on zero-shot 3D classification. For each test shape, we extract its 3D feature and match it with text embeddings of candidate class names, assigning the most similar one as the predicted label. Besides OmniObject3D, we also report classification accuracy on ModelNet40~\cite{modelnet40} test set (2,468 shapes in 40 classes) under canonical and randomly rotated conditions.

We can see from Tab.~\ref{tab:zeroshot_cls} that non-RI models perform well under canonical poses but their accuracy drops with random rotations. Among RI methods, RI-Transformer achieves the highest accuracy on ModelNet40, while RI-Mamba surpasses other baselines on the more challenging OmniObject3D. This may be because ModelNet40 contains only 40 categories, many of which overlap with the pretraining data, allowing the attention layers in RI-Transformer to overfit. In contrast, RI-Mamba exhibits stronger generalization to unseen categories, highlighting its balanced representational power and robustness for real-world classification.

\begin{table}[h]
\centering
    \Huge 
    \resizebox{0.45\textwidth}{!}{%
    \begin{tabular}{lccccc}
    \toprule
    Data            & Method         & MNet40 & MNet40-SO(3) & Omni3D & Omni3D-SO(3) \\
    \midrule
    \midrule    
    \multirow{5}{*}{\rotatebox{90}{ShapeNet}}        & PointBERT-TAMM & \textbf{73.2}       & 16.9           & \underline{15.3}         & 5.1              \\
                    & DuoMamba      &     72.6       &       16.4         & \textbf{15.4}              &       4.3           \\ 
                    \cline{2-6}\rule{0pt}{2.2ex}
                    & LocoTrans      &      66.9      &       66.8         &       11.3       &       11.3           \\ 
                    & RITransformer  & \underline{72.9}       & \textbf{72.7}           & 13.2        & \underline{13.4}             \\
                    & RI-Mamba (ours)       & 71.9       & \underline{71.8}           & {14.7}         & \textbf{14.2}             \\  
    \midrule
    \multirow{5}{*}{\rotatebox{90}{Ensemble}}        & PointBERT-TAMM & \textbf{81.1}       & 25.3           & \underline{34.0}           & 16.5             \\
                    & DuoMamba      &    \underline{80.5}        &       24.8         &    32.9          &        17.0          \\
                    \cline{2-6}\rule{0pt}{2.2ex}
                    & LocoTrans      &     73.1       &      73.0          &      21.6        &       21.4           \\ 
                    & RITransformer  & 78.3       & \textbf{78.7}           & 30.0           & \underline{30.6}             \\
                    & RI-Mamba (ours)       & 76.7       & \underline{76.5}           & \textbf{34.4}         & \textbf{34.5}     \\ 
    \bottomrule
    \end{tabular}
    }
\caption{Zero-shot classification accuracy on ModelNet40 and OmniObject3D.}
\label{tab:zeroshot_cls}
\end{table}

\subsection{Ablation Study}
\paragraph{Component Contribution.} 
We analyze the impact of each component in RI-Mamba, including Hilbert sorting, reference frame disambiguation (RFD), positional and orientational embeddings, feature-wise linear modulation (FiLM), and bidirectional scanning in Tab.~\ref{tab:component}. Removing any component degrades performance, indicating that each plays a distinct and complementary role in RI-Mamba. Notably, removing positional and orientational embeddings (row \texttt{(3)}) leads to a substantial performance drop from 14.7 to 8.2. Reintegrating them without adaptive FiLM partially recovers performance but remains suboptimal (row \texttt{(4)}). Without bidirectional scanning, performance reduces to 12.7 (last row). These results highlight the critical role of bidirectional information flow across point patches and the adaptive integration of geometric context via FiLM.
\begin{table}[t]
\centering
    \resizebox{0.43\textwidth}{!}{%
    \begin{tabular}{ccc|ccc|c|c} 
    \toprule
    & Hilbert & RFD & $\mathbf{pos}$ & $\mathbf{ori}$ & FiLM & BiScan & Omni3D \\  
    \midrule
    \midrule
    \texttt{(0)} & \cmark & \cmark & \cmark & \cmark & \cmark & \cmark & \textbf{14.7}  \\ 
    \midrule    
    \texttt{(1)} & - & \cmark & \cmark & \cmark & \cmark & \cmark & 13.9 \\ 
    \texttt{(2)} & \cmark & - & \cmark & \cmark & \cmark & \cmark & 14.1  \\ 
    \midrule
    \texttt{(3)} & \cmark & \cmark & - & -  & - & \cmark & 8.2  \\ 
    \texttt{(4)} & \cmark & \cmark & \cmark & \cmark & - & \cmark & 13.1 \\ 
    \midrule
    \texttt{(5)} & \cmark & \cmark & \cmark & \cmark & \cmark & - & 12.7  \\  
    \bottomrule
    \end{tabular}    
    }
\caption{Contribution of each component in RI-Mamba.}
\label{tab:component}
\end{table}

\begin{figure}[t]
    \centering
    \hspace{-5mm}
    \begin{subfigure}[t]{0.16\textwidth}
        \centering
        \includegraphics[height=2.4cm]{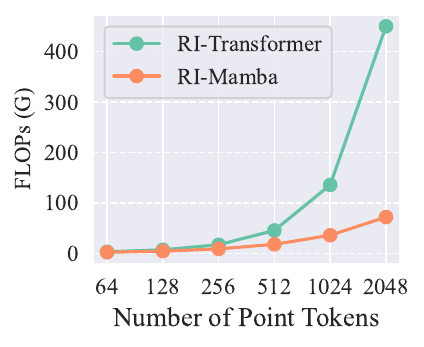}
    \end{subfigure} 
    \begin{subfigure}[t]{0.16\textwidth}
        \centering
        \includegraphics[height=2.4cm]{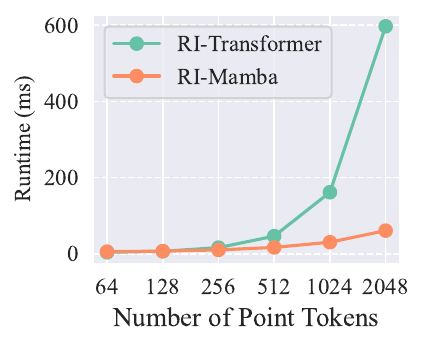}
    \end{subfigure} 
    \begin{subfigure}[t]{0.16\textwidth}
        \centering
        \includegraphics[height=2.4cm]{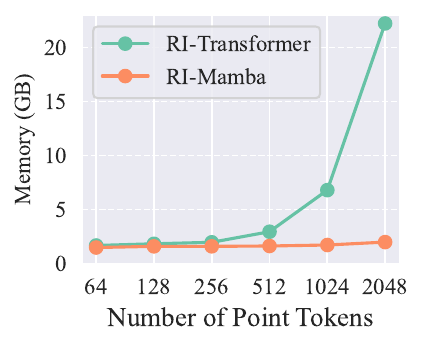}
    \end{subfigure}    
    \caption{RI-Mamba and RI-Transformer efficiency comparisons.} 
    \label{fig:inference_efficiency}
\vspace{-3mm}    
\end{figure}

\paragraph{Computational Efficiency.} In Fig.~\ref{fig:inference_efficiency}, we compare the inference efficiency of RI-Mamba and RI-Transformer, measured by FLOPs, runtime, and memory usage.
As the number of point tokens increases, the required FLOPs, memory, and runtime of RI-Transformer grow rapidly, while RI-Mamba remains considerably lower and scales more efficiently. In particular, the GPU memory required by RI-Transformer exceeds 20 GB at a token length of 2048, whereas RI-Mamba maintains a memory usage of around 2 GB. This efficiency is attributed to our linear patch-wise orientational embedding and state-space model design, making RI-Mamba more practical and scalable for real-world deployment.

\begin{figure}[t]
    \centering
    \hspace{-5mm}
    \begin{subfigure}[t]{0.26\textwidth}
        \centering
        \includegraphics[height=2.5cm]{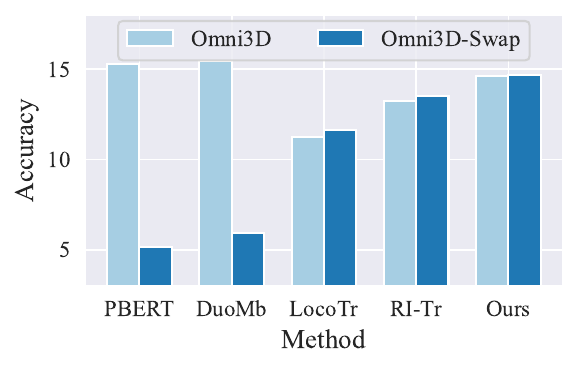}
        \caption{Performance on $y$-$z$ axis swap.}
        \label{fig:axis_swap}
    \end{subfigure}\hspace{-2mm}
    \begin{subfigure}[t]{0.22\textwidth}
        \centering
        \includegraphics[height=2.5cm]{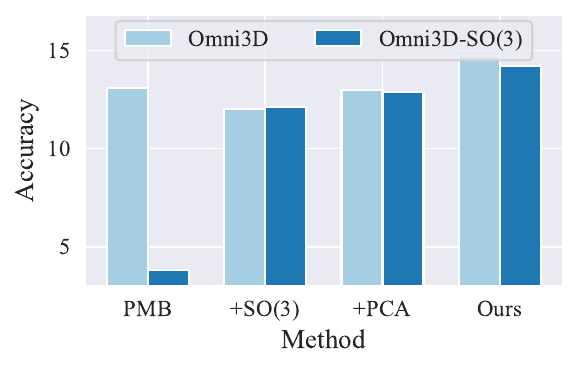}
        \caption{Alternative RI methods.}
        \label{fig:alternative_ri}
    \end{subfigure}    
    \caption{Comparison on axis swap robustness and alternative RI approaches.} 
    \label{fig:axis_wap_other_ri}
\vspace{-3mm}    
\end{figure}

\paragraph{Robustness to Axis Swap.} 
We evaluate the robustness of different models to gravity-axis swapping in Fig.~\ref{fig:axis_swap}. This variation commonly occurs in real-world scenarios, as datasets often adopt inconsistent conventions for the gravity axis (either $y$ or $z$), posing challenges for stable feature extraction. As shown, non-RI methods such as PointBERT and DuoMamba exhibit high sensitivity to axis swapping, leading to substantial performance degradation. In contrast, RI-based models demonstrate strong robustness, with our RI-Mamba achieving the highest performance.

\paragraph{Alternative RI Approaches for Mamba.} 
We explore alternative strategies to introduce rotation invariance into PointMamba \cite{pointmamba}, a state-space point cloud encoder that closely resembles our model. We evaluate two methods: (1) augmenting training data with SO(3) rotations, and (2) applying PCA-based alignment to normalize the input point clouds. As shown in Fig.~\ref{fig:alternative_ri}, the vanilla PointMamba (PMB) performs poorly on rotated inputs. Incorporating random rotations during training improves robustness at test time, but significantly reduces performance on canonical data.  
PCA-based alignment offers more stable performance across both settings, though it slightly underperforms compared to the original PMB on canonical data. In contrast, our RI-Mamba consistently achieves the highest accuracy on both aligned and rotated point clouds, benefiting from carefully designed components that preserve geometric information typically lost in RI pipelines while maintaining high model expressiveness.

\section{Conclusion}
In this paper, we presented a robust solution for text-to-shape retrieval.
At the core of our solution is RI-Mamba, the first rotation-invariant architecture based on state-space models, capable of extracting robust shape embeddings with linear-time complexity. We further adopt cross-modal contrastive learning with automatically generated triplets to eliminate the need for manual annotations and achieve scalable training. For the first time, our method enables text-to-shape retrieval on over 200 object categories with arbitrary orientations, closely reflecting real-world scenarios.

\paragraph{Limitations.} Despite its robustness, RI-Mamba relies on PCA-based reference frames to achieve rotation invariance, making it susceptible to the inherent instability of PCA. This can lead to sensitivity on symmetric or near-symmetric shapes and to noise in point clouds. A deeper analysis of these limitations and the exploration of more stable mechanisms will be an important direction for future work. 
\section{Acknowledgments}
This research was supported by the Australian Government through the Australian Research Council's Discovery Projects funding scheme (project DP240101926). Professor Ajmal Mian is the recipient of an ARC  Future Fellowship Award (project FT210100268) funded by the Australian Government.

{
    \small
    \bibliographystyle{ieeenat_fullname}
    \bibliography{main}
}

\clearpage
\setcounter{page}{1}
\maketitlesupplementary

\section{Training and Implementation Details}
We implement our method in PyTorch~\cite{pytorch} and conduct all experiments on a single NVIDIA RTX 4090 GPU with 24GB memory. Our encoder consists of 12 RI-Mamba blocks with a feature dimension of 512. For RI positional and orientational embeddings, we use a two-layer MLP with a hidden size of 128 and GELU activation~\cite{gelu}. The FiLM module includes a bottleneck reducing the feature dimension from 512 to 128, followed by a two-layer MLP (hidden size 128) to compute the modulation parameters $\gamma$ and $\beta$.

For pretraining, we adopt the cross-modal contrastive learning framework introduced in TAMM, training the 3D encoders for 200 epochs with 10 warm-up epochs. We use the AdamW optimizer \cite{adamw} and a cosine learning rate schedule with a base learning rate of $5\mathrm{e}{-4}$. The CLIP image adapter from TAMM is used to reduce the domain gap between rendered and real images. All experiments use the OpenCLIP-ViT-G/14 variant \cite{openclip} as pre-trained image-text encoders. For patch-based encoders (PointBERT, DuoMamba, RI-Transformer, RI-Mamba), the input point cloud consists of 10K points. For the graph-based encoder LocoTrans, we use 1024 points following the setting in their paper. We follow original configurations and set the number of point patches ($G$) to 384 for PointBERT, 256 for DuoMamba, and 64 for RI-Mamba and RI-Transformer. The number of neighbors ($k$) is set to 64 for PointBERT, 20 for LocoTrans, and 32 for RI-Mamba and RI-Transformer.

Results of PointBERT pretrained on ShapeNet in Tab.~\ref{tab:zeroshot_text2shape}-\ref{tab:zeroshot_cls} are obtained using the released weights from TAMM. For all other zero-shot results, we pretrain the encoders using the configuration described above. All results are reported from a single run with a fixed random seed of 0. For the supervised text-to-shape experiment (Tab.~\ref{tab:text2shape}), RI-Mamba is trained with the same setup. To evaluate on Text2Shape-SO(3) (Tab.~\ref{tab:text2shape}), we directly use the released weights of SCA3D, while Parts2Words is trained using the official implementation and default hyperparameters due to the lack of available pretrained weights.

\section{OmniObject3D Captions}
For each object in the OmniObject3D dataset \cite{omniobject3d}, a detailed textual annotation is provided, including a high-level summary as well as descriptions of appearance, material, style, and function, as illustrated in examples 1 and 2 below.
To construct text queries for zero-shot text-to-shape retrieval (Sec.~\ref{subsec:zeroshot_text2shape}), we concatenate the \textbf{Summary} and \textbf{Appearance} fields. This ensures that each query captures both semantic class information (from the summary) and fine-grained visual details (from the appearance description), enabling better discrimination among objects belonging to the same category.

\begin{tcolorbox}[title=OmniObject3D: Example 1, colback=gray!5, colframe=black!40]
\label{box:example1}
\small
\textbf{Summary:} A dark green table.

\textbf{Appearance:} This is a rectangular table, made of a flatter rectangle for the tabletop, the rest of the rectangle is sealed, made of bamboo, with square edges for the skeleton, the overall color is dark green, left and right symmetrical.

\textbf{Material:} Bamboo, hard, slightly reflective, rough surface.

\textbf{Style:} Realistic.

\textbf{Function:} Placing items to assist with work.

\vspace{0.5em}
\begin{tcolorbox}[colback=yellow!15, colframe=yellow!40!black, title=\textbf{Constructed Query}, left=1mm, right=1mm]
A dark green table. This is a rectangular table, made of a flatter rectangle for the tabletop, the rest of the rectangle is sealed, made of bamboo, with square edges for the skeleton, the overall color is dark green, left and right symmetrical.
\end{tcolorbox}

\end{tcolorbox}

\begin{tcolorbox}[title=OmniObject3D: Example 2, colback=gray!5, colframe=black!40]
\label{box:example2}
\small
\textbf{Summary:} It's a toy train.

\textbf{Appearance:} This toy train as a whole is green, the top part is yellowish green, there is a round bump, the lower part of the front and back of each side of a cylindrical bump, this train has a total of four wheels, the train surface does not have any hand-painted patterns, the overall structure of the axisymmetric.

\textbf{Material:} Plastic, rubber, iron, hard, smooth surface, slightly reflective.

\textbf{Style:} Cartoon.

\textbf{Function:} Entertainment, decoration.

\vspace{0.5em}
\begin{tcolorbox}[colback=yellow!15, colframe=yellow!40!black, title=\textbf{Constructed Query}, left=1mm, right=1mm]
It's a toy train. This toy train as a whole is green, the top part is yellowish green, there is a round bump, the lower part of the front and back of each side of a cylindrical bump, this train has a total of four wheels, the train surface does not have any hand-painted patterns, the overall structure of the axisymmetric.
\end{tcolorbox}

\end{tcolorbox}

\onecolumn
\section{Additional Visualization on Text2Shape}
\begin{figure*}[h!]
    \centering
    \includegraphics[width=0.9\linewidth]{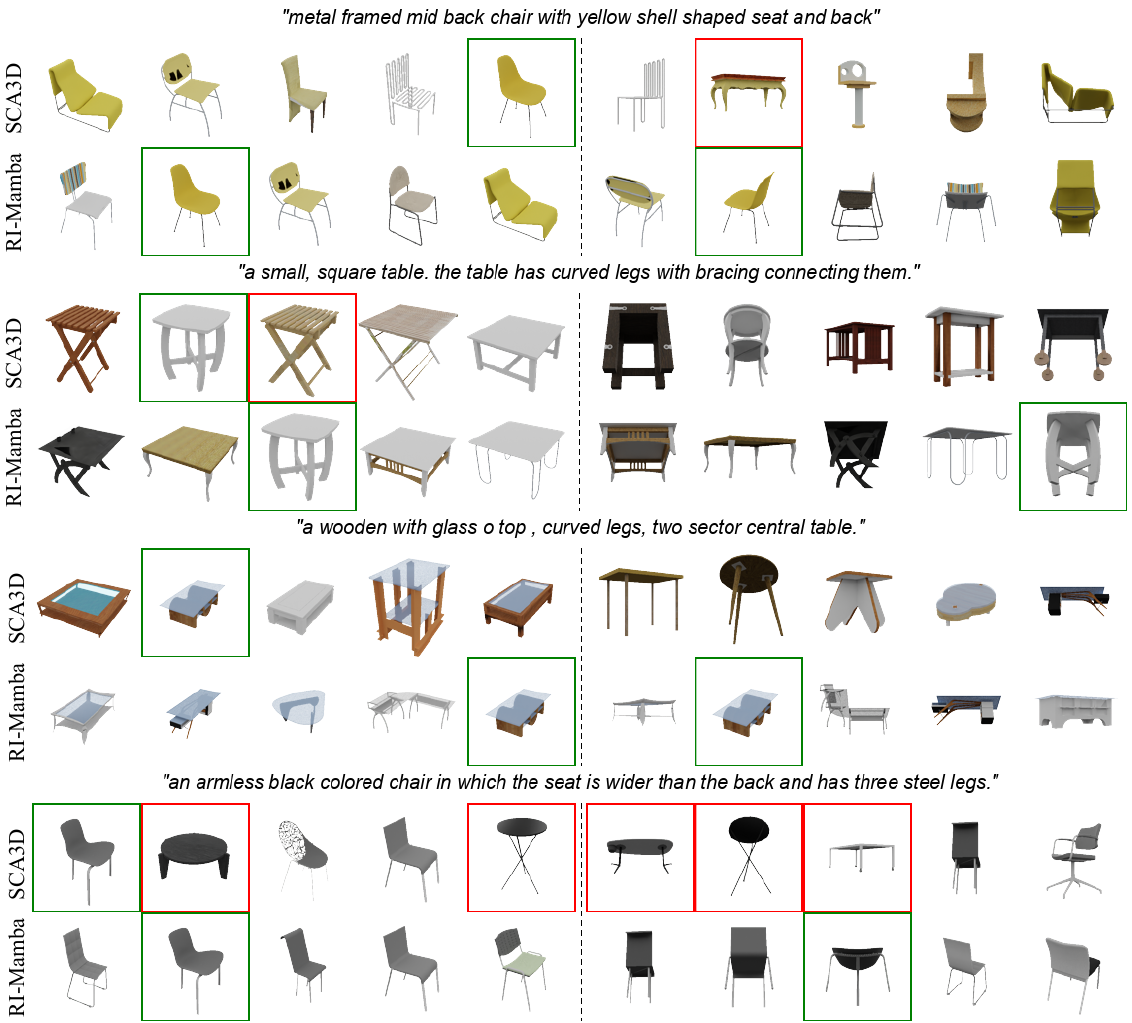}
    \caption{Retrieval results of SCA3D and RI-Mamba on the Text2Shape dataset. The target object is highlighted in the green box, and objects from the incorrect class are marked with red boxes.}
    \label{fig:supp_text2pcd_retrieval_match_a}
\end{figure*}

\begin{figure*}[t]
    \centering
    \includegraphics[width=0.9\linewidth]{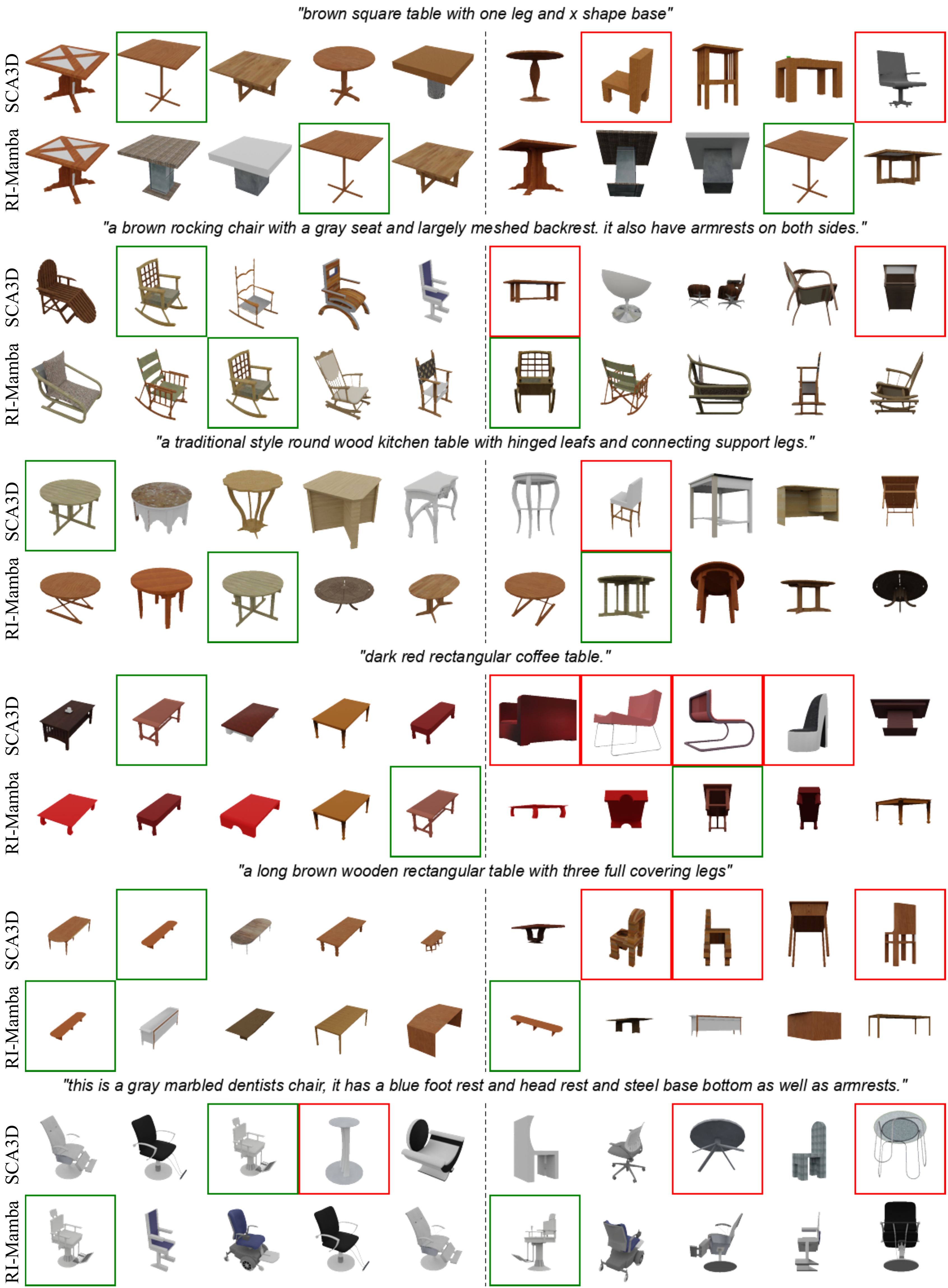}
    \caption{Retrieval results of SCA3D and RI-Mamba on the Text2Shape dataset. The target object is highlighted in the green box, and objects from the incorrect class are marked with red boxes (continued).}
    \label{fig:supp_text2pcd_retrieval_match5_b}
\end{figure*}

\begin{figure*}[h]
    \centering
    \includegraphics[width=0.9\linewidth]{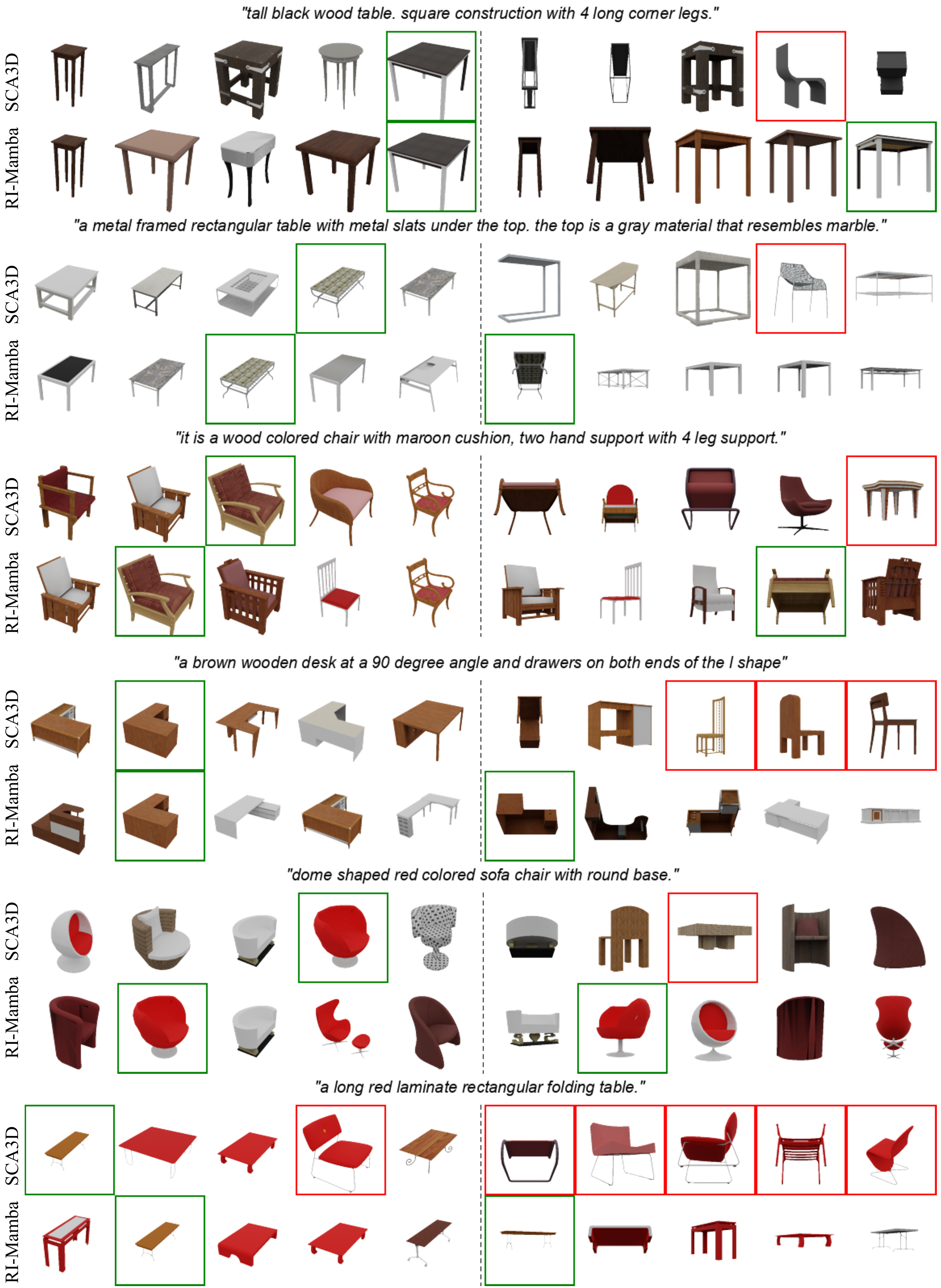}
    \caption{Retrieval results of SCA3D and RI-Mamba on the Text2Shape dataset. The target object is highlighted in the green box, and objects from the incorrect class are marked with red boxes (continued).}
    \label{fig:supp_text2pcd_retrieval_match4_a}
\end{figure*}

We provide extra visualization for text-to-shape retrieval results of SCA3D and RI-Mamba on the Text2Shape benchmark in Fig.~\ref{fig:supp_text2pcd_retrieval_match_a}, Fig.~\ref{fig:supp_text2pcd_retrieval_match5_b}, and Fig.~\ref{fig:supp_text2pcd_retrieval_match4_a}. As we can see, RI-Mamba demonstrates strong robustness to rotations, whereas SCA3D performs well only under canonical poses. Under rotation, SCA3D struggles to distinguish tables and chairs, often failing to retrieve the correct object described in the text. These results underscore the challenges of real-world retrieval scenarios, where objects appear in varied orientations, and highlight the practical effectiveness of our proposed RI-Mamba.

\end{document}